\documentclass[nohyperref]{article}

\usepackage{microtype}
\usepackage{graphicx}
\usepackage{subfigure}
\usepackage{booktabs} 

\usepackage{hyperref}

\usepackage[accepted]{icml2022}

\usepackage{amsmath}
\usepackage{amssymb}
\usepackage{mathtools}
\usepackage{amsthm}
\usepackage{multirow}
\usepackage{gensymb}
\usepackage[capitalize,noabbrev]{cleveref}

\theoremstyle{plain}

\theoremstyle{definition}

\theoremstyle{remark}

\newcommand{\Mat}{\boldsymbol}
\newcommand{\Set}{\mathcal}

\newcommand{\real}{\mathbb{R}}

\DeclareMathOperator{\Var}{Var}
\DeclareMathOperator*{\argmin}{arg\,min}

\usepackage[textsize=tiny]{todonotes}

\icmltitlerunning{Neural Implicit Dictionary Learning via Mixture-of-Expert Training}

\begin{document}

\twocolumn[
\icmltitle{Neural Implicit Dictionary Learning via Mixture-of-Expert Training}




\begin{icmlauthorlist}
\icmlauthor{Peihao Wang}{uta}
\icmlauthor{Zhiwen Fan}{uta}
\icmlauthor{Tianlong Chen}{uta}
\icmlauthor{Zhangyang Wang}{uta}
\end{icmlauthorlist}

\icmlaffiliation{uta}{Department of Electrical and Computer Engineering, University of Texas at Austin}

\icmlcorrespondingauthor{Zhangyang Wang}{atlaswang@utexas.edu}

\icmlkeywords{Machine Learning, ICML}

\vskip 0.3in
]



\printAffiliationsAndNotice{}  

\begin{abstract}
Representing visual signals by coordinate-based deep fully-connected networks has been shown advantageous in fitting complex details and solving inverse problems than discrete grid-based representation.
However, acquiring such a continuous Implicit Neural Representation (INR) requires tedious per-scene training on tons of signal measurements, which limits its practicality.
In this paper, we present a generic INR framework that achieves both data and training efficiency by learning a Neural Implicit Dictionary (NID) from a data collection and representing INR as a functional combination of basis sampled from the dictionary.
Our NID assembles a group of coordinate-based subnetworks which are tuned to span the desired function space.
After training, one can instantly and robustly acquire an unseen scene representation by solving the coding coefficients.
To parallelly optimize a large group of networks, we borrow the idea from Mixture-of-Expert (MoE) to design and train our network with a sparse gating mechanism.
Our experiments show that, NID can improve reconstruction of 2D images or 3D scenes by 2 orders of magnitude faster with up to 98\% less input data.
We further demonstrate various applications of NID in image inpainting and occlusion removal, which are considered to be challenging with vanilla INR.
Our codes are available in \url{https://github.com/VITA-Group/Neural-Implicit-Dict}.
\vspace{-5mm}
\end{abstract}

\section{Introduction}
\label{sec:intro}

Implicit Neural Representations (INRs) have recently demonstrated remarkable performance in representing multimedia signals in computer vision and graphics \citep{park2019deepsdf, mescheder2019occupancy, saito2019pifu, chen2021learning, sitzmann2020implicit, tancik2020fourier, mildenhall2020nerf}.
In contrast to classical discrete representations, where real-world signals are sampled and vectorized before processing, INR directly parameterizes the continuous mapping between coordinates and signal values using deep fully-connected networks (also known as multi-layer perceptron or MLP).
This continuous parameterization enables to represent more complex and flexible scenes without being limited by grid extents and resolution in a more compact and memory efficient way.

However, one significant drawback of this approach is that acquiring an INR usually requires a \textit{tedious} \textit{per-scene} training of neural networks on \textit{dense} measurements, which limits the practicality.
\citet{yu2021pixelnerf, wang2021ibrnet, chen2021mvsnerf} generalizes Neural Radiance Field (NeRF) \citep{mildenhall2020nerf} across various scenes by projecting image features to a 3D volumetric proxy and then rendering feature volume to generate novel views.
To speed up INR training, \citet{sitzmann2020metasdf, tancik2021learned} apply meta-learning algorithms to learn the initial weight parameters for the MLP based on the underlying class of signals being represented.
However, this line of works are either hard to be extended beyond NeRF scenario or incapable of producing high-fidelity results with insufficient supervision.

In this paper, we design a unified INR framework that simultaneously achieves optimization and data efficiency.
We think of reconstructing an INR from few-shot measurements as solving an underdetermined system.
Inspired by compressed sensing techniques \citep{donoho2006compressed}, we represent every neural implicit function as a linear combination of a function basis sampled from an over-complete \textit{Neural Implicit Dictionary} (\textbf{NID}).
Unlike conventional basis representation as a wide matrix, an NID is parameterized by a group of small neural networks that acts as continuous function basis spanning the entire target function space.
The NID is shared across different scenes while the sparse codes are specified by each scene.
We first acquire the NID ``offline" by jointly optimizing it with per-scene coding across a class of instances in a training set.
When transferring to unseen scenarios, we re-use the NID and only solves the the scene specific coding coefficients ``online".

To effectively scale to thousands of subnetworks inside our dictionary, we employ the Mixture-of-Expert (MoE) training for NID learning  \citep{shazeer2017outrageously}.
We model each function basis in our dictionary as an expert subnetwork and the coding coefficients as its gating state. During each feed-forward, we utilize a routing module to generate sparsely coded gates, i.e., activating a handful of basis experts and linearly combining their responses. Training with MoE also ``kills two birds with one stone" by constructing transferable dictionaries and avoiding extra computational overheads.

Our contributions can be summarized as follows:
\begin{itemize}
\item We propose a novel data-driven framework to learn a \textit{Neural Implicit Dictionary} (\textbf{NID}) that can transfer across scenes, to both accelerate per-scene neural encoding and boost their performance. 
\item NID is parameterized by a group of small neural networks that acts as continuous function basis to span the neural implicit function space. The dictionary learning is efficiently accomplished via MoE training. 
\item We conduct extensive experiments to validate the effectiveness of NID. 
For \underline{training efficiency}, we show that our approach is able to achieve 100$\times$ faster convergence speed for image regression task.
For \underline{data efficiency}, our NID can reconstruct signed distance function with 98\% less point samples, and optimize a CT image with 90\% fewer views. We also demonstrate more practical applications for NID, including image inpainting, medical image recovery, and transient object detection for surveillance videos.
\end{itemize}

\section{Preliminaries}
\label{sec:related_work}

\paragraph{Compressed Sensing in Inverse Imaging.} \label{sec:cps}
Compressed sensing and dictionary learning are widely applied in inverse imaging problems~\cite{lustig2008compressed,metzler2016denoising,fan2018segmentation}. In classical signal processing, signals are discretized and represented by vectors.
A common goal is to reconstruct signals (or digital images) $\Mat{x} \in \real^{N}$ from $M$ measurements $\Mat{y} \in \real^M$, which are formed by linearly transforming the underlying signals plus noise: $\Mat{y} = \Mat{A}\Mat{x} + \Mat{\eta}$.
However, $\Mat{A}$ is often highly ill-posed, \textit{i.e.}, number of measurements is much smaller than the number of unknowns ($M \ll N$), which makes this inverse problem rather challenging.
Compressed sensing \citep{candes2006robust, donoho2006compressed} provides an efficient approach to solve this underdetermined linear system by assuming signals $\Mat{x} \in \real^{N}$ are compressible and representing it in terms of few vectors inside a group of spanning vectors $\Mat{\Psi} = \begin{bmatrix} \Mat{\psi}_i & \cdots & \Mat{\psi}_{K} \end{bmatrix} \in \real^{N \times K}$.
Then we can reconstruct $\Mat{x}$ through the following optimization objective:
\begin{align} \label{eqn:csp}
    \argmin_{\Mat{\alpha}} \lVert\Mat{\alpha}\rVert_0 \text{ subject to } \left\lVert \Mat{y} - \Mat{A} \Mat{\Psi} \Mat{\alpha} \right\rVert_2 < \varepsilon
\end{align}
where $\Mat{\alpha} \in \real^{K}$ is known as the sparse code coefficient, and $\lVert \Mat{\eta} \rVert_2 \le \varepsilon$ is a bound on the noise level. One often  replaces the $\ell_0$ semi-norm with $\ell_1$ to obtain a convex objective.
The spanning vectors $\Mat{\Psi}$ can be chosen from orthonormal bases or, more often than not, over-complete dictionaries ($N \ll K$) \cite{kreutz2003dictionary, tovsic2011dictionary, aharon2006k, chen2016compressed}.
Rather than a bunch of spanning vectors, \citet{chan2015pcanet, tariyal2016deep, papyan2017convolutional} proposed hierarchical dictionary implemented by neural network layers.

\paragraph{Implicit Neural Representation.} \label{sec:inr}
Implicit Neural Representation (INR) in computer vision and graphics replaces traditional discrete representations of multimedia objects with continuous functions parameterized by multilayer perceptrons (MLP) \citep{tancik2020fourier, sitzmann2020implicit}.
Since this representation is amenable to gradient-based optimization, prior works managed to apply coordinate-based MLPs to many inverse problems in computational photography \citep{park2019deepsdf, mescheder2019occupancy, mildenhall2020nerf, chen2021learning, chen2021nerv, sitzmann2021light, fan2022unified, attal2021torf, shen2021non} and scientific computing \citep{han2018solving, li2020fourier, zhong2021cryodrgn}.
Formally, we denote an INR inside a function space $\Set{F}$ by $f_{\theta}: \real^m \rightarrow \real$, which continuously maps $m$-dimension spatio-temporal coordinates (say $(x, y)$ with $m = 2$ for images) to the value space (say pixel intensity). Consider a functional $\mathcal{R}: \Set{F} \times \Omega \rightarrow \real$, we intend to find the network weights $\theta^*$ such that:
\begin{align}
    \mathcal{R}(f_{\theta^*} \vert \omega) = 0, \text{for every } \omega \in \Omega
\end{align}
where $\Omega$ records the measurement settings. For instance, in computed tomography (CT), $\mathcal{R}$ is called the volumetric projection integral and $\Omega$ specifies the ray parameterization and corresponding colors. When solving ordinal differential equations, $\mathcal{R}$ takes form of $\rho(\Mat{x}, f, \nabla f, \nabla^2 f, ...)$ if $\Mat{x} \in \Omega\setminus\partial\Omega$, while $\mathcal{R} = f(\Mat{x}) - C$ for some constant $C$ if $\Mat{x} \in \partial\Omega$, given a compact set $\Omega$ and operator $\rho(\cdot)$ which combines derivatives of $f$ \citep{sitzmann2020implicit}.

\paragraph{Mixture-of-Expert Training.} \label{sec:moe}
\citet{shazeer2017outrageously} proposed outrageously wide neural networks with dynamic routing to achieve larger model capacity and higher data parallel.
Their approach is to introduce an Mixture-of-Expert (MoE) layer with a number of expert subnetworks and train a gating network to select a sparse combination of the experts to process each input.
Let us denote by $G(\Mat{x})$ and $E_i(\Mat{x})$ the output of the gating network and the output of the $i$-th expert network for a given input $\Mat{x}$. The output of the MoE module can be written as:
\begin{align} \label{eqn:gate}
    y = \sum_{i=1}^n G(\Mat{x})_i E_i(\Mat{x}),
\end{align}
where $n$ is the number of experts and $\lVert G(\Mat{x}) \rVert_0 = k$. In \citet{shazeer2017outrageously}, computation is saved based on the sparsity of $G(\Mat{x})$. The common sparsification strategy is called noisy top-$k$ gating, which can be formulated as:
\begin{align}
& G(\Mat{x}) = \operatorname{Normalize}(\operatorname{TopK}(H(\Mat{x}), k)), \\
& \operatorname{TopK}(\Mat{x}, k)_i = \left\{\begin{array}{ll}
x_i & \text{if $x_i$ is in top $k$ elements} \\
0 & \text{otherwise}
\end{array}\right.,
\end{align}
where $H(\Mat{x})$ synthesizes raw gating activations, $\operatorname{TopK}(\cdot)$ masks out $n-k$ smallest elements, and $\operatorname{Normalize}(\cdot)$ scales the magnitude of remaining weights to a constant, which can be chosen from softmax or $\ell_p$-norm normalization.

\section{Neural Implicit Dictionary Learning}

As we discussed before, inverse imaging problems are often ill-posed and it is also true for Implicit Neural Representation (INR).
Moreover, training an INR network is also time-consuming.
How to kill two bird with one stone by efficiently and robustly acquiring an INR from few-shot observations remains uninvestigated.
In this section, we answer this question by presenting our approach \textit{Neural Implicit Dictionary} (NID), which are learned from data collections \textit{a priori} and can be re-used to quickly fit an INR.
We will first reinterpret two-layer SIREN \citep{sitzmann2020implicit} and point out the limitation of current design.
Then we will elaborate on our proposed models and the techniques to improve its generalizability and stability.

\subsection{Motivation by Two-Layer SIREN} \label{sec:motivation}

Common INR architectures are pure Multi-Layer Perceptrons (MLP) with periodic activation functions.
Fourier Feature Mapping (FFM) \citep{tancik2020fourier} places a sinusoidal transformation  after the first linear layer, while Sinusoidal Representation Network (SIREN) \citep{sitzmann2020implicit} replaces every nonlinear activation with a sinusoidal function.
For the sake of simplicity, we only consider two-layer INR architectures to unify the formulation of FFM and SIREN.
To be consistent with the notation in Section \ref{sec:inr}, let us denote INR by function $f: \real^m \rightarrow \real$, which can be formulated as below:
\begin{align}
    \label{eqn:pe} & \gamma(\Mat{x}) = \begin{bmatrix} \sin(\Mat{w}_1^T \Mat{x} + b_1) & \cdots & \sin(\Mat{w}_n^T \Mat{x} + b_n) \end{bmatrix}^T, \\
    \label{eqn:siren_lin} & f(\Mat{x}) = \Mat{\alpha}^T \gamma(\Mat{x}) + c,
\end{align}
where $\Mat{w}_i \in \real^m, b_i \in \real, \forall i \in [n]$ and $\Mat{\alpha} \in \real^n, c \in \real$ are all network parameters, and mapping $\gamma(\cdot)$ (cf. Equation \ref{eqn:pe}) is called positional embedding \cite{mildenhall2020nerf, zhong2021cryodrgn}.
After simply rewriting, we can obtain:
\begin{align}
    f(\Mat{x}) &= \sum_{i=1}^n \alpha_i \sin(\Mat{w}_i^T \Mat{x} + b_i) + c \\
    \label{eqn:hartley} &\approx \int_{\real^m} \frac{\alpha(\Mat{w})}{\sqrt{\pi}}  \sin\left(\Mat{w}^T \Mat{x} + \frac{\pi}{4} \right) \mathrm{d}\Mat{\Mat{w}},
\end{align}
from which we discover Equations \ref{eqn:pe}-\ref{eqn:siren_lin} can be considered as an approximation of inverse Hartley (Fourier) transform (cf. Equation \ref{eqn:hartley}).
The weights of the first SIREN layer sample frequency bands on the Fourier domain, and  
passing coordinates through sinusoidal activation functions maps spatial positions onto cosine-sine wavelets.
Then training a two-layer SIREN amounts to finding the optimal frequency supports and fitting the coefficients in Hartley transform.

Although trigonometric polynomials are dense in continuous function space, cosine-sine waves may not be always desirable as approximating functions at arbitrary precision with finite neurons can be infeasible.
In fact, some other bases, such as Gegenbauer basis \cite{feng2021signet} and Pl\"ucker embedding \cite{attal2021learning},  have been proven useful in different tasks.
However, we argue that since handcrafted bases are agnostic to data distribution, they cannot express intrinsic information about data, thus may generalize poorly across various scenes.
This causes per-scene training to re-select the frequency supports and re-fit the Fourier coefficients.
Moreover, when observations are scarce, sinusoidal basis can also result in severe over-fitting in reconstruction \citep{sutherland2015error}.

\begin{figure}
\centering
\includegraphics[width=0.96\linewidth]{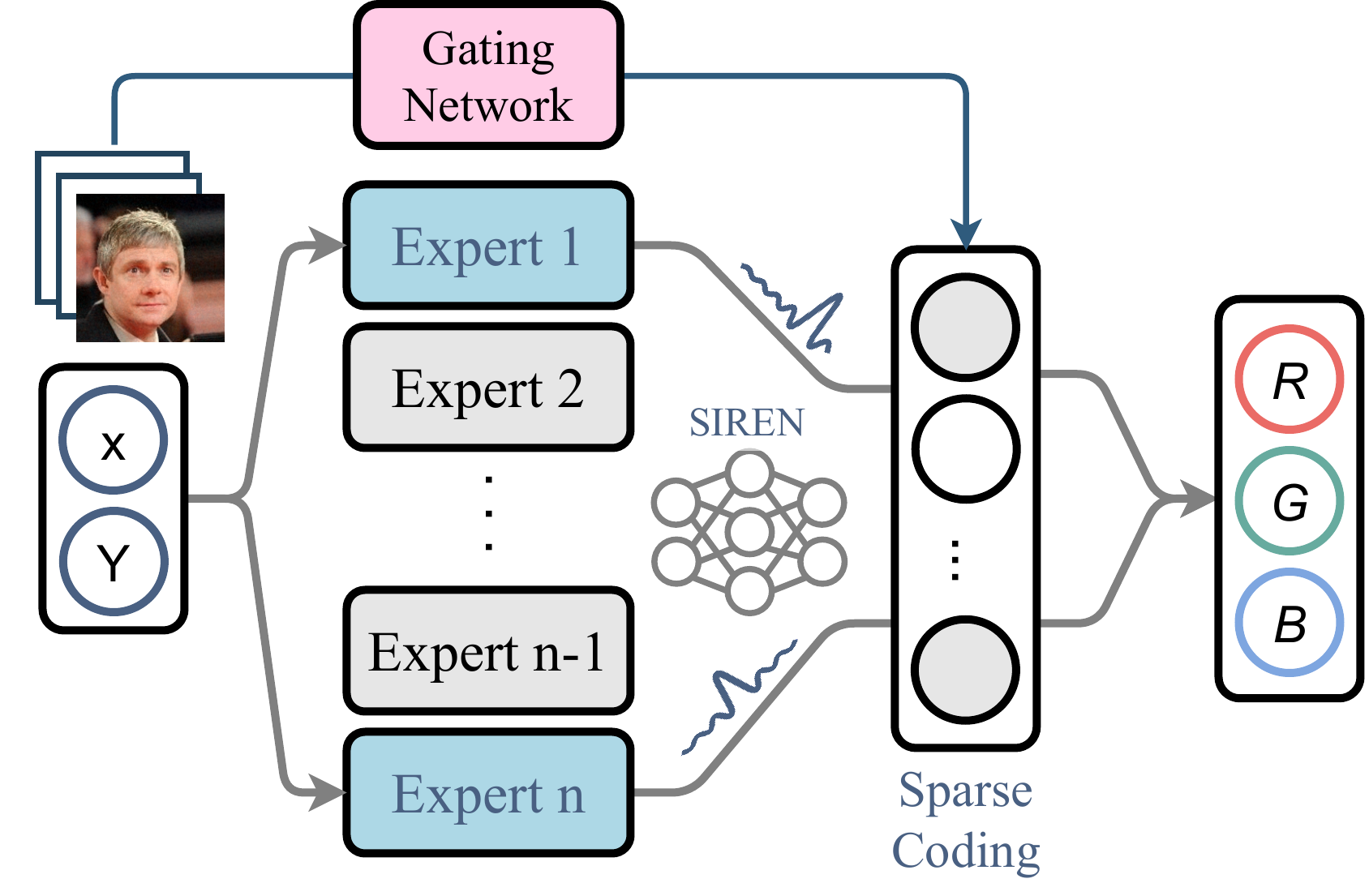}
\vspace{-1em}
\caption{Illustration of our NID pipeline. The blue experts are activated while grey ones are ignored.}
\label{fig:pipeline}
\vspace{-1em}
\end{figure}

\subsection{Learning Implicit Function Basis} \label{sec:learn_basis}

Having reasoned why current INR architectures generalize badly and demand tons of measurements, we intend to introduce the philosophy of sparse dictionary representation \cite{kreutz2003dictionary, tovsic2011dictionary, aharon2006k} into INR.
A dictionary contains a group of over-complete basis that spans the signal space.
In contrast to handcrafted bases or wavelets, dictionary are usually learned from a data collection.
Since it is aware of the distribution of the underlying signals to be represented, expressing signals using dictionary enjoys higher sparsity, robustness and generalization power.

Even though dictionary learning algorithms are well established in \citet{aharon2006k}, it is far from trivial to design dictionaries amenable to INR on the continuous domain.
Formally, we want to obtain a set of continuous maps: $b_i: \real^m \rightarrow \real, \forall i \in [n]$ such that for every signal $f: \real^m \rightarrow \real$ inside our target signal space $\Set{F}$, there exists a sparse coding $\Mat{\alpha} \in \real^{n}$ that can express the signal:
\begin{align}
f(\Mat{x}) = \alpha_1 b_1(\Mat{x}) + \cdots + \alpha_n b_n(\Mat{x}) \quad \forall \Mat{x} \in \real^m,
\end{align}
where $n$ is the size of the dictionary, and $\Mat{\alpha}$ satisfies $\lVert \Mat{\alpha} \rVert_0 \le k$ for some sparsity $k \ll n$.
We parameterize each component in the dictionary with small coordinate-based networks by $b_{\theta_1}, \cdots, b_{\theta_n}$, where $\theta_i$ denotes the network weights of the $i$-th element.
We call this group of function basis \textit{Neural Implicit Dictionary (NID)}.

We adopt an end-to-end optimization scheme to learn the NID. During training stage, we jointly optimize the subnetworks inside NID and the sparse coding assigned with each instance.
Suppose we own a data collection with measurements captured from $T$ multimedia instances to be represented (say $T$ images or geometries of objects): $\Set{D} = \{ \Mat{\Omega}^{(i)} \in \real^{t_i \times m}, \Mat{Y}^{(i)} \in \real^{t_i} \}_{i=1}^{T}$, where $\Mat{\Omega}^{(i)}$ is the observation parameters (say coordinates on 2D lattice for images), $m$ is the dimension of such parameters, $\Mat{Y}^{(i)}$ are measured observations (say corresponding RGB colors), $t_i$ denotes the number of observations for $i$-th instance.
Then we optimize the following objective on the training dataset:
\begin{align} \label{eqn:objective}
    & \argmin_{\theta_1, \cdots, \theta_n \atop \Mat{\alpha}^{(1)}, \cdots, \Mat{\alpha}^{(T)}} \sum_{i=1}^{T} \sum_{j=1}^{t_i} \mathcal{L}\left(\mathcal{R}(f^{(i)} \vert \Mat{\Omega}^{(i)}_j), \Mat{Y}^{(i)}_{j}\right) \\
    &\quad\quad\quad\quad\quad + \lambda \mathcal{P}\left(\Mat{\alpha}^{(i)}, \cdots, \Mat{\alpha}^{(T)}\right), \nonumber \\
    & \text{subject to } f^{(i)}(\Mat{x}) = \sum_{j=1}^{n} \alpha^{(i)}_j b_{\theta_j}(\Mat{x}) \quad \forall \Mat{x} \in \real^m, \nonumber
\end{align}
where $f^{(i)} \in \Set{F}$ is the INR of the $i$-th instance, $\mathcal{R}(f \vert \Mat{\omega}): \Set{F} \times \Mat{\Omega} \rightarrow \real$ is a functional measuring function $f$ with respect to a group of parameters $\Mat{\omega}$.
$\mathcal{L}(\cdot)$ is the loss function dependent of downstream tasks.
$\mathcal{P}(\cdot)$ places a regularization onto the sparse coding, $\lambda = 0.01$ is fixed in our experiments. Besides sparsity penalty, we also consider some joint prior distributions among all codings, which will be discussed in Section \ref{sec:train_moe}.
When transferring to unseen scenes, we fix NID basis $\{b_{\theta_i}\}_{i=1}^N$ and only compute the corresponding sparse coding to minimize the objective in Equation \ref{eqn:objective}.


\subsection{Training Thousands of Subnetworks with Mixture-of-Expert Layer} \label{sec:train_moe}

Directly invoking thousands of networks causes inefficiency and redundancy due to sample dependent sparsity.
Moreover, this brute force computational strategy fails to properly utilize the advantage of modern computing architectures in parallelism.
As we introduced in Section \ref{sec:moe}, Mixture-of-Expert (MoE) training system \citep{shazeer2017outrageously, he2021fastmoe} provides a conditional computation mechanism that achieves stable and parallel training on a outrageously large networks.
We notice that MoE layer and NID share the intrinsic similarity in the underlying running paradigm.
Therefore, we propose to leverage an MoE layer to represent an NID accommodating thousands of implicit function basis.
Specifically, each element in NID is an expert network in MoE layer, and the sparse coding encodes the gating states.
Below we elaborate on the implementation details of the MoE based NID layer part by part:

\paragraph{Expert Networks.}
Each expert network is a small SIREN \citep{sitzmann2020implicit} or FFM \citep{tancik2020fourier} network.
To downsize the whole MoE layer, we share the positional embedding and the first 4 layers among all expert networks.
Then we append two independent layers for each expert.
We note this design can make two experts share the early-stage features and adjust their coherence.

\paragraph{Gating Networks.}
The generated gating is used as the sparse coding of an INR instance.
We provide two alternatives to obtain the gating values:
1) We employ an encoder network as the gating function to map the (partial) observed measurements to the pre-sparsified weights. For  grid-like modality, we utilize convolutional neural networks (CNN) \citep{he2016deep, liu2018image, gordon2019convolutional}. For unstructured point modality, we adopt set encoders \citep{zaheer2017deep, qi2017pointnet, qi2017pointnet++}.
2) We can also leverage a lookup table \citep{bojanowski2017optimizing} where each scene is assigned with a trainable embedding jointly optimized with expert networks.
After computing the raw gating weights, we recall the method in Equation \ref{eqn:gate} to sparsify gates.
Different from \citet{shazeer2017outrageously}, we do not perform softmax normalization to gating logits. Instead, we sort gating weights with respect to their absolute values, and normalize the weights by its $\ell_2$ norm.
Comparing aforementioned two gating functions, encoder-based gating networks benefit in parameter saving and instant inference without need of re-fitting sparse coding.
However, headless embeddings demonstrate more strength in training efficiency and achieve better convergence.

\begin{table*}[t]
\caption{Performance of NID compared with FFM, SIREN, and Meta on CelebA dataset. $\uparrow$ the higher the better, $\downarrow$ the lower the better. The unit of \# Params is megabytes, FLOPs is in gigabytes, and throughput is in \#images/s. }
\label{tbl:img_results}
\vskip 0.15in
\begin{center}
\resizebox{0.8\linewidth}{!}{
\begin{tabular}{l|ccc|ccc}
\toprule
Methods & PSNR ($\uparrow$) & SSIM ($\uparrow$) & LPIPS ($\downarrow$) & \# Params & FLOPs & Throughput \\
\midrule
FFM \citep{tancik2020fourier} & 22.60 & 0.636 & 0.244 & 147.8 & 20.87 & 0.479 \\
SIREN \citep{sitzmann2020implicit} & 26.11 & 0.758 & 0.379 & 66.56 & 4.217 & 0.540 \\
\midrule
Meta + 5 steps \citep{tancik2021learned} & 23.92 & 0.583 & 0.322 & 66.69 & 4.217 & 0.536 \\
Meta + 10 steps \citep{tancik2021learned} & 29.64 & 0.651 & 0.182 & 66.69 & 4.217 & 0.536 \\
\midrule
NID + init. ($k=128$) & 28.75 & 0.892 & 0.061 & 8.972 & 23.30 & 30.37 \\
NID + 5 steps ($k=128$) & 33.57 & 0.941 & 0.027 & 8.972 & 23.30 & 30.37 \\
NID + 10 steps ($k=128$) & 35.10 & 0.954 & 0.021 & 8.972 & 23.30 & 30.37 \\
\midrule
NID + init. ($k=256$) & 30.26 & 0.919 & 0.045 & 8.972 & 29.55 & 21.23 \\
NID + 5 steps ($k=256$) & 35.09 & 0.960 & 0.019 & 8.972 & 29.55 & 21.23 \\
NID + 10 steps ($k=256$) & 37.75 & 0.971 & 0.012 & 8.972 & 29.55 & 21.23 \\
\bottomrule
\end{tabular}
}
\end{center}
\vskip -0.1in
\end{table*}

\paragraph{Patch-wise Dictionary.}
It is implausible to construct an over-complete dictionary to represent entire signals.
We adopt the walkround in \cite{reiser2021kilonerf, turki2021mega} by partitioning the coordinate space into regular and overlapped patches, and assign separate NID to each block.
We implement this by setting up multiple MoE layers and dispatch the coordinate inputs to corresponding MoE with respect to the region where they are located.

\paragraph{Utilization Balancing and Warm-Up.}
It was observed that gating network tends to converge to a self-reinforcing imbalanced state, where it always produces large weights for the same few experts \citep{shazeer2017outrageously}.
To tackle this problem, we pose a regularization on the Coefficient of Variation (CV) of the sparse codings following \citet{bengio2015conditional, shazeer2017outrageously}.
The CV penalty is defined as:
\begin{align}
    & \mathcal{P}_{CV}\left(\Mat{\alpha}^{(i)}, \cdots, \Mat{\alpha}^{(T)}\right) = \frac{\Var(\Mat{\bar{\alpha}})}{\left(\sum_{i=1}^{n}\Mat{\bar{\alpha}}_i / n\right)^2}, \\
    & \text{where } \Mat{\bar{\alpha}} = \sum_{i=1}^{T} \Mat{\alpha}^{(i)}.
\end{align}
Evaluating this regularization over the whole training set is infeasible. Instead we estimate and minimize this loss per batch.
We also find hard sparsification will stop gradient back-propagation, which leads to stationary gating states equal to the initial stage.
To address this side-effect, we first abandon hard thresholding and train the MoE layer with $\ell_1$ penalty $\mathcal{P}_{\ell_1} = \sum_{i=1}^{T} \lVert\Mat{\alpha}^{(i)}\rVert_1$ on codings for several epochs, and enable sparsification afterwards.

\begin{figure}[t]
\begin{center}
\includegraphics[width=1\linewidth]{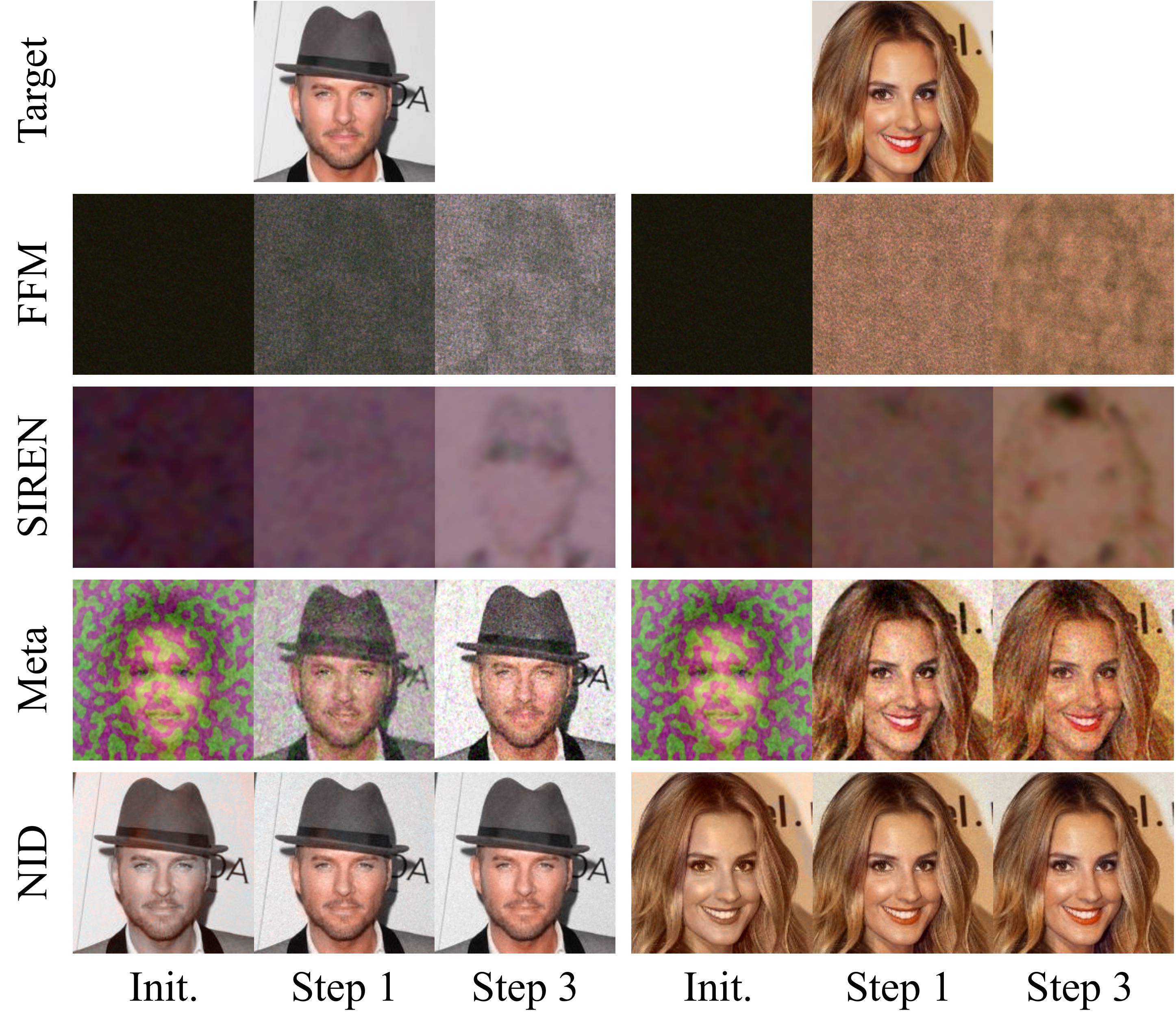}
\vspace{-2em}
\caption{A closer look at the early training stages of FFM, SIREN, Meta, and NID, respectively.}
\label{fig:image_res}
\end{center}
\vspace{-2em}
\end{figure}

\section{Experiments and Applications}

In this section, we demonstrate the promise of NID by showing several applications in scene representation. 
\subsection{Instant Image Regression} \label{sec:img_regress}

A prototypical example of INR is to regress a 2D image with an MLP which takes in coordinates on 2D lattice and is supervised with RGB colors.
Given a $D \times D$ image $\Mat{Y} \in \real^{D \times D \times 3}$, our goal is to approximate the mapping $f: \real^2 \mapsto \real^3$ by optimizing $\lVert f(i, j) - \Mat{Y}_{ij} \rVert_2$ for every $(i, j) \in [0, D]^2$, where $f_\theta = \sum_i \alpha_i b_{\theta_i}$.
In conventional training scheme, each image is encoded into a dedicated network after thousands of iterations.
Instead, we intend to use NID to instantly acquire such INR without training or with only few steps of gradient descent.

\begin{figure*}[h]
\begin{center}
\includegraphics[width=0.95\linewidth]{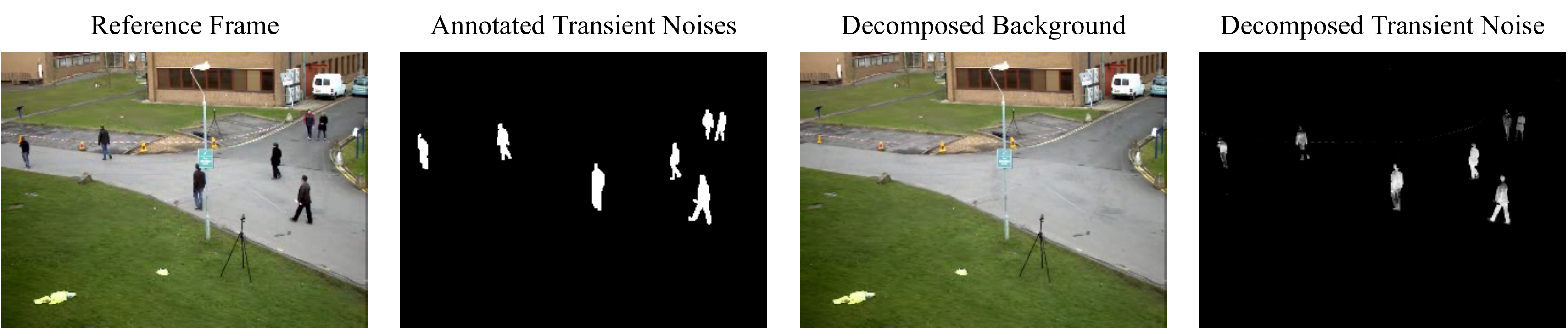}
\caption{Visualization of foreground-background decomposition results for surveillance video via principal component pursuit with NID.}
\label{fig:video_res}
\end{center}
\end{figure*}

\paragraph{Experimental Settings.}
We choose to train our NID on CelebA face dataset \citep{liu2015faceattributes}, where each image is cropped to $178 \times 178$.
Our NID contains 4096 experts, each of which share a 4-layer backbone with 256 hidden dimension and own a separate 32-dimension output layer.
We adopt 4 residual convolutional blocks \citep{he2016deep} as the gating network. 
During training, the gating network is tuned with the dictionary.
NID is warmed up within 10 epochs and then start to only keep top 128 experts for each input for 5000 epochs.
At the inference stage, we let gating network directly output the sparse coding of the test image.
To further improve the precision, we utilize the output as the initialization, and then use gradient descent to further optimize the sparse coding with the dictionary fixed.
We contrast our methods to FFM \citep{tancik2020fourier}, SIREN \citep{sitzmann2020implicit} and Meta \citep{tancik2021learned}.
In Table \ref{tbl:img_results}, we demonstrate the overall PSRN, SSIM \citep{wang2004image}, and LPIPS \citep{zhang2018unreasonable} of these four models on test set (with 500 images) under the limited training step setting, where FFM and SIREN are only trained for 100 steps.
We also present the inference time metrics in Table \ref{tbl:img_results}, including the number of parameters to represent 500 images, FLOPs to render a single image, and measured throughput of images rendered per second.
In Figure \ref{fig:image_res}, we zoom into the initialization and early training stages of each model.

\paragraph{Results.}
Results in Table \ref{tbl:img_results} show that NID ($k=256$) can achieve best performance among all compared models even without subsequent optimization steps.
A relative sparser NID ($k=128$) can also surpass both FFM and SIREN (trained with 100 steps) with the initially inferred coding.
Compared with meta-learning based method, our model can outperform them by a significant margin ($\ge 5\mathrm{dB}$) within the same optimization steps.
We note that since NID only further tunes the coding vector, both computation and convergence speed are much faster than meta-learning approaches which fine-tune parameters of the whole network.
Figure \ref{fig:image_res} illustrates that the initial sparse coding inferred from the gating network is enough to produce high-accuracy reconstructed images.
With 3 more gradient descent steps (which usually takes 5 seconds), it can reach the quality of well-tuned per-scene training INR (which takes 10 minutes).
We argue that although meta learning is able to find a reasonable start point, but the subsequent optimization is sensitive to saddle points where the represented images are fuzzy and noisy.
In regard to model efficiency, our NID is 8 times more compact than single-MLP representation, as NID shares dictionary among all samples and only needs to additionally record an small gating network.
Moreover, our MoE implementation results in a significant throughput gain, as it makes inference highly parallelable.
We point out that meta-learning can only provide an initialization. To represent all test images, one has to save all dense parameters separately.
Horizontally compared, denser NID is more expressive than sparser one though sacrificing efficiency.

\subsection{Facial Image Inpainting.} \label{sec:img_inpaint}
 
\begin{figure}[t]
\vskip 0.2in
\begin{center}
\includegraphics[width=0.99\linewidth]{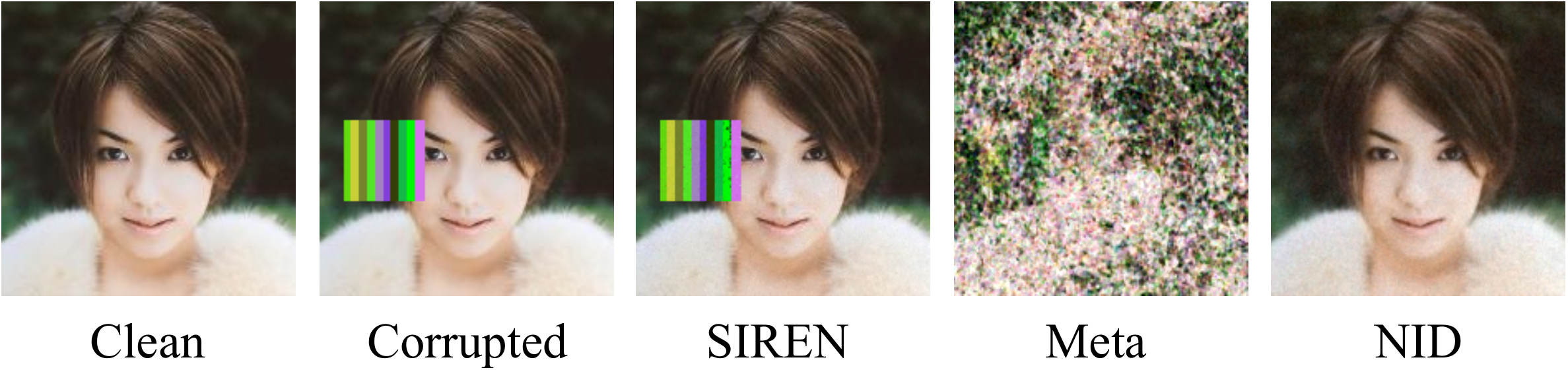}
\caption{Qualitative results of inpainting image from corruptions with NID.}
\label{fig:image_occ}
\end{center}
\vskip -0.2in
\end{figure}

Image inpainting recovers images corrupted by occlusion. Previous works \citep{liu2018image, yu2019free} only establish algorithms based on discrete representation. In this section, we demonstrate image inpainting directly on continuous INR.
Given a corrupted image $\Mat{Y} \in \real^{D \times D \times 3}$, we remove outliers by projecting $\Mat{Y}$ onto some low-dimension linear (function) subspace spanned by components in a dictionary.
We achieve this by trying to represent the corrupted image as a linear combination of a pre-trained NID, while simultaneously enforcing the sparsity of this combination. Specifically, we fix the dictionary in Equation \ref{eqn:objective} and choose $\ell_1$ norm as the loss function $\mathcal{L}$ \citep{candes2011robust}:
\begin{align}
    \label{eqn:inpainting_objective}
    & \argmin_{\Mat{\alpha} \in \real^n} \sum_{(x,y) \in [0, D]^2} \left\lVert \sum_{i=1}^{n} \alpha_i b_{\theta_i}(x, y) - \Mat{Y}_{xy}\right\rVert_1, \\
    & \text{subject to } \lVert\Mat{\alpha}\rVert_0 \le k, \nonumber
\end{align}
where we assume noises are sparsely distributed on images.

\paragraph{Experimental Settings.}
We corrupt images by randomly pasting a $48 \times 48$ color patch.
To recover images, we borrow the dictionary trained on CelebA dataset from Section \ref{sec:img_regress}.
However, we do not leverage the gating network to synthesize the sparse coding. Instead, we directly optimize a randomly initialized coding to minimize Equation \ref{eqn:objective}.
Our baseline includes SIREN and Meta \citep{tancik2021learned}.
We change their loss function to $\ell_1$ norm to keep consistent.
To inpaint with Meta, we start from its learned initialization, and optimize two steps towards the objective.

\paragraph{Results.}
The inpainting results are presented in Figure \ref{fig:image_occ}.
Our findings are 1) SIREN overfits all given signals as it does not rely on any image priors.
2) Meta-learning based approach implicitly poses a prior by initializing the networks around a desirable optimum. However, our experiment shows that the learned initialization is ad-hoc to a certain data distribution. When noises are added, 
Meta turns unstable and converges to a trivial solution.
3) Our NID displays stronger robustness by accurately locating and removing the occlusion pattern.

\subsection{Self-Supervised Surveillance Video Analysis}

In this section, we establish a self-supervision algorithm that can decompose foreground and background for surveillance videos based on NID.
Given a set of video frames $\{\Mat{Y}^{(t)} \in \real^{D \times D \times 3} \}_{t=1}^{T}$, our goal is to find a continuous mapping $f(x,y,t)$ representing the clip that can be decomposed to: $f(x,y,t) = f_X(x,y,t) + f_E(x,y,t)$, where $f_X$ is the background and $f_E$ are transient noises (\textit{e.g.}, pedestrians).
We borrow the idea from Robust Principal Component Analysis (RPCA) \citep{candes2011robust, ji2010robust} where background is assumed to be ``low-rank" and noises are assumed to be sparse.
Despite well-established for discrete representation, modeling ``low-rank" in continuous domain remains elusive.
We achieve this by assuming $f_X(x,y,t)$ at each time stamp are largely represented by the same group of experts, \textit{i.e.}, the non-zero elements in the sparse codings concentrate to several points, and the coding weights follow a decay distribution.
Mathematically, we first rewrite $f$ by decoupling spatial coordinates and time: $f(x, y, t) = \sum_{i} \alpha_i(t) b_{\theta_i}(x, y)$, where every time slice shares a same dictionary, and sparse coding $\alpha_i(t)$ depends on the timestamp.
Then we minimize:
\begin{align}
    \label{eqn:rpca_objective}
    & \argmin_{\theta_1, \cdots, \theta_n, \atop \alpha(t)} \sum_{t=1}^{T} \sum_{(x,y)} \left\lVert \sum_{i=1}^n \alpha_i(t) b_{\theta_i}(x, y) - \Mat{Y}^{(t)}_{xy}\right\rVert_1 \\
    &\quad\quad\quad\quad\quad + \lambda \sum_{t=1}^{T} \sum_{i=1}^{n} \frac{\lvert\alpha_{i}(t)\rvert}{\exp(-\beta i)}, \nonumber
\end{align}
where the second term penalize the sparsity of $\alpha(t)$ according to an exponentially increasing curve (controlled by $\beta$), which implies the larger $i$ is, the more sparsity is enforced.
As a consequence, every time slice are largely approximated by the first few components in NID, which simulates the nature of ``low-rank" representation for continuous functions.

\paragraph{Results.}
We test the above algorithm on BMC-Real dataset \citep{vacavant2012benchmark}.
In our implementation, $\alpha(t)$ is also parameterized by another MLP, and we choose $\beta = 0.5$.
Our qualitative results are presented in Figure \ref{fig:video_res}.
We verify that our algorithm can decompose the background and foreground correctly by imitating the behavior of RPCA.
This application further demonstrates the potential of our NID in combining with subspace learning techniques.

\begin{table}[h]
\caption{Quantitative results of CT reconstruction compared with FFM, SIREN, and Meta. (PSNR in dB)}
\label{tbl:ct_res}
\begin{center}
\resizebox{0.48\textwidth}{!}{
\begin{tabular}{l|cccccc}
\toprule
\multirow{2}{*}{Methods} & \multicolumn{2}{c}{128 views} & \multicolumn{2}{c}{16 views}  & \multicolumn{2}{c}{8 views}\\
& PSNR & SSIM & PSNR & SSIM & PSNR & SSIM \\
\midrule
FFM \citep{tancik2020fourier} & 22.81 & 0.845 & 15.22 & 0.122 & 13.58 & 0.095 \\
SIREN \citep{sitzmann2020implicit} & 24.32 & 0.891 & 18.48 & 0.510 & 17.26 & 0.483 \\
Meta \citep{tancik2021learned} & 32.70 & \textbf{0.948} & 21.39 & 0.822 & \textbf{18.28} & 0.574 \\
\midrule
NID ($k=128$) & 36.56 & 0.939 & 24.48 & 0.818 & 16.24 & 0.619 \\
NID ($k=256$) & \textbf{37.49} & 0.944 & \textbf{26.32} & \textbf{0.829} & 16.77 & \textbf{0.636}\\
\bottomrule
\end{tabular}}
\vspace{-2em}
\end{center}
\end{table}

\begin{figure}[t!]
\begin{center}
\includegraphics[width=0.90\linewidth]{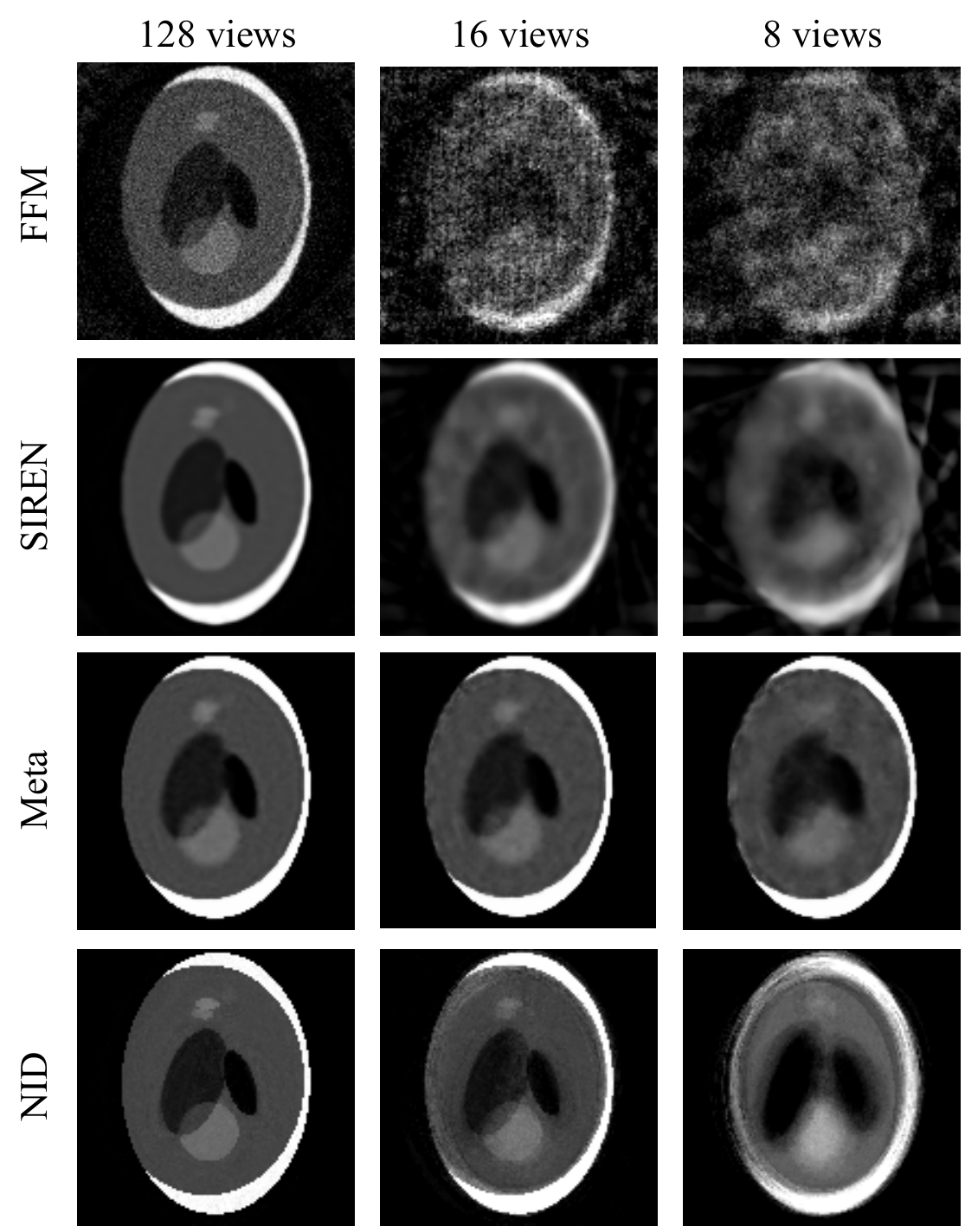}
\caption{Qualitative results of CT reconstruction from sparse measurements.}
\label{fig:ct_res}
\end{center}
\vspace{-2em}
\end{figure}

\subsection{Computed Tomography Reconstruction}

Computed tomography (CT) is a widely used medical imaging technique that captures projective measurements of the volumetric density of body tissue.
This imaging formation can be formulated as below:
\begin{align}
Y(r, \phi) = \int_{\real^2} f(x, y) \delta(r - x \cos\phi - y\sin\phi) \mathrm{d}x\mathrm{d}y,
\end{align}
where $r$ is the location on the image plane, $\phi$ is the viewing angle, and $\delta(\cdot)$ is known as Dirac delta function. 
Due to limited number of measurements, reconstructing $f$ through inversing this integral is often ill-posed.
We propose to shrink the solution space by using NID as a regularization.

\paragraph{Experimental Settings.}
We conduct experiments on Shepp-Logan phantoms dataset \citep{shepp1974fourier} with 2048 randomly generated $128 \times 128$ CTs.
We first directly train an NID over 1k CT images, during which the total number of experts is 1024, and each CT selects 128/256 experts. 
In CT scenario, a look-up table is chosen as our gating network.
Afterwards, we randomly sample 128 viewing angles, and synthesize 2D integral projections of a bundle of 128 parallel rays from these angles as the measurement.
To testify the effectiveness of our method under limited number of observations, we downsample 128 views by 12.5\%(16) and 6.25\%(8) respectively.
Again, we choose FFM \citep{tancik2020fourier}, SIREN \citep{sitzmann2020implicit}, and Meta \citep{tancik2021learned} as our baselines.

\paragraph{Results.}
The quantitative results are listed in Table \ref{tbl:ct_res}.
We observe that our NID consistently leads two metrics in the table. When sampled views are sufficient, NID achieves the highest PSNR, while when views are reduced, our NID takes advantage in SSIM.
We also plot the qualitative results in Figure \ref{fig:ct_res}. We find that our NID can regularize the reconstructed results to be smooth and shape-consistent, which leads to less missing wedge artifacts.

\subsection{Shape Representation from Point Clouds}

Recent works \citep{park2019deepsdf, sitzmann2020metasdf, sitzmann2020implicit, gropp2020implicit} convert point clouds to continuous surface representation through directly regressing a Signed Distance Function (SDF) parameterized by MLPs.
Suppose $f: \real^3 \rightarrow \real$ is our target SDF, given a set of points $\Omega \subset \real^3$, we fit $f$ by solving a integral equation of the form below \citep{park2019deepsdf}:
\begin{align} \label{eqn:sdf_objective}
\argmin_{f} & \int_{\Mat{x} \in \Omega}  \lvert f(\Mat{x}) \rvert \mathrm{d}\Mat{x}
+ \int_{\Mat{x} \in \real^3\setminus\Omega} \lvert f(\Mat{x}) - d(\Mat{x}, \Omega) \rvert \mathrm{d}\Mat{x},
\end{align}
where $d(\Mat{x}, \Omega)$ denotes the signed shortest distance from point $\Mat{x}$ to point set $\Omega$.
During optimization, we evaluate the first integral via sampling inside the given point cloud and the second term via uniformly sampling over the whole space.
Tackling this integral with sparsely sampled points around the surface is challenging \citep{park2019deepsdf}.
Similarly, we introduce NID to learn \textit{a priori} SDF basis from data and then leverage it to regularize the solution.

\begin{table}[h]
\vspace{-1em}
\caption{Quantitative results of SDF reconstruction compared with SIREN, DeepSDF, MetaSDF. CD is short for Chamfer Distance (magnified by $10^3$), NC means Normal Consistency. $\uparrow$ the higher the better, $\downarrow$ the lower the better. }
\label{tbl:sdf}
\begin{center}
\resizebox{0.48\textwidth}{!}{
\begin{tabular}{l|cccccc}
\toprule
\multirow{2}{*}{Methods} & \multicolumn{2}{c}{500k points} & \multicolumn{2}{c}{50k points}  & \multicolumn{2}{c}{10k points}\\
& CD($\downarrow$) & NC($\uparrow$) & CD($\downarrow$) & NC($\uparrow$) & CD($\downarrow$) & NC($\uparrow$) \\
\midrule
SIREN \citep{sitzmann2020implicit} & \textbf{0.051} & \textbf{0.962} & 0.163 & 0.801 & 1.304 & 0.169 \\
IGR \citep{gropp2020implicit} & 0.062 & 0.927 & 0.170 & 0.812 & 0.961 & 0.676 \\
DeepSDF \citep{park2019deepsdf} & 0.059 & 0.925  & 0.121 & 0.856 & 2.751 & 0.194  \\
MetaSDF \citep{sitzmann2020metasdf} & 0.067 & 0.884 & 0.097 & 0.878 & 0.132 & 0.755 \\
ConvONet \citep{peng2020convolutional} & 0.052 & 0.938 & 0.082 & 0.914 & 0.133 & 0.845 \\
\midrule
NID ($k=128$) & 0.058 & 0.940 & 0.067 & 0.948 & 0.093 & 0.921 \\
NID ($k=256$) & 0.053 & 0.956 & \textbf{0.063} & \textbf{0.952} & \textbf{0.088} & \textbf{0.945} \\
\bottomrule
\end{tabular}}
\vspace{-1em}
\end{center}
\end{table}

\paragraph{Experimental Settings.}
Our experiments about SDF are conducted on ShapeNet \citep{shapenet2015} datasets, from which we pick the \textit{chair} category for demonstration.
To guarantee meshes are watertight, we run the toolkit provided by \citet{huang2018robust} to convert the whole dataset.
We split the \textit{chair} category following \citet{choy20163d}, and fit our NID over the training set.
The total number of experts is 4096, and after 20 warm-up epochs, only 128/256 experts will be preserved for each sample.
We choose lookup table as our gating network.
During inference time, we sample 500k, 50k and 10k point clouds, respectively, from the test surfaces.
Then we optimize objective in Equation \ref{eqn:sdf_objective} to obtain the regressed SDF with $f$ represented by our NID.
In addition to SIREN and IGR \cite{gropp2020implicit}, We choose DeepSDF \citep{park2019deepsdf}, MetaSDF \citep{sitzmann2020metasdf}, and ConvONet \citep{peng2020convolutional} as our baselines.
Our evaluation metrics are Chamfer distance (the average minimal pairwise distance) and normal consistency (the angle between corresponding normals).

\begin{figure}[h]
\vspace{-1em}
\begin{center}
\includegraphics[width=0.90\linewidth]{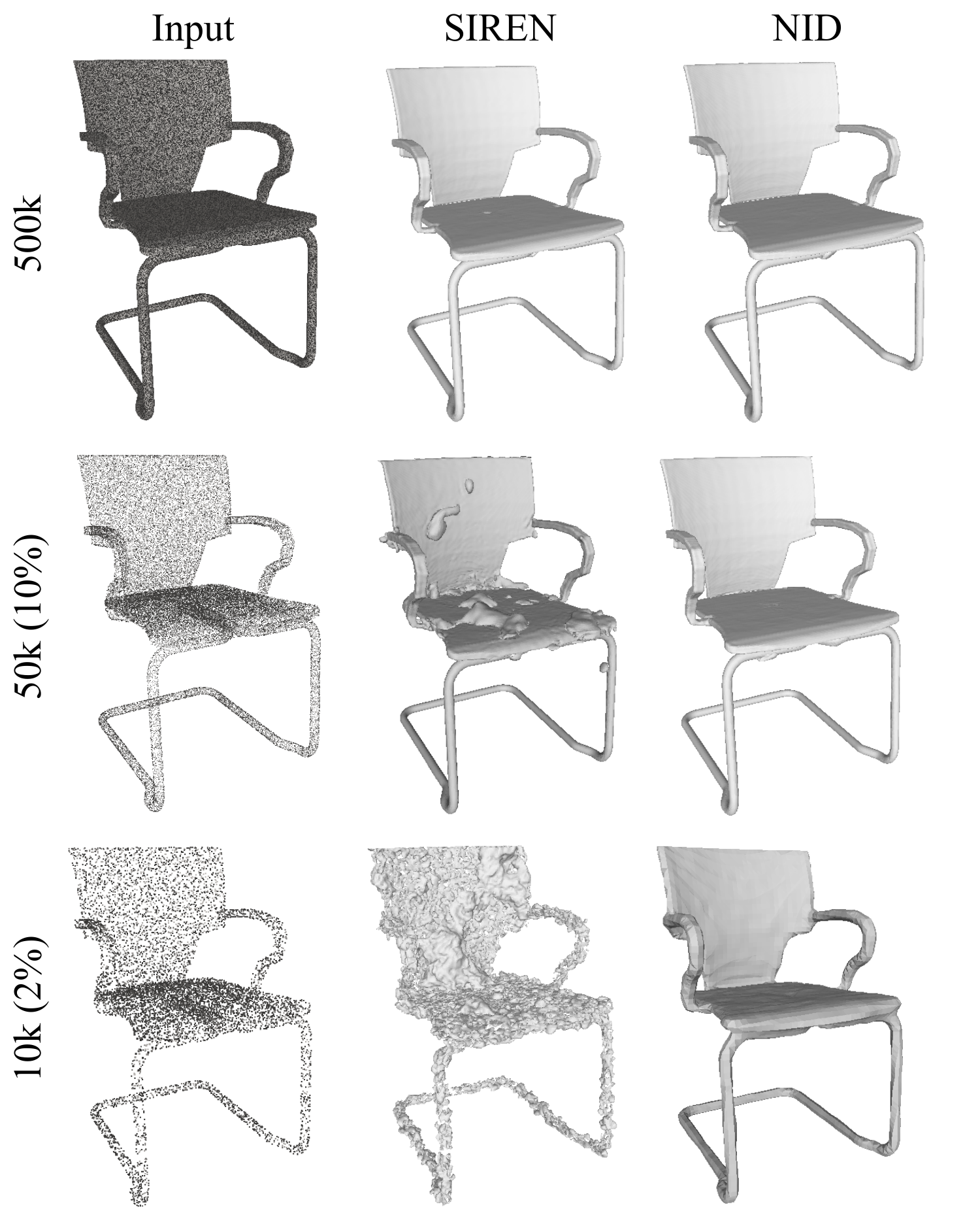}
\caption{Qualitative results of SDF reconstruction from sparse point clouds.}
\label{fig:sdf_res}
\vspace{-2em}
\end{center}
\end{figure}

\paragraph{Results.}
We put our numerical results in Table \ref{tbl:sdf}, from which we can summarize that our NID is more robust to smaller number of points. As the performance of other methods drops quickly, the CD metric of NID stays below 0.1 and NC keeps above 0.9.
We also provide qualitative illustration in Figure \ref{fig:sdf_res}.
We conclude that thanks to the constraint of our NID, the SDF will not collapse at some point where observations are missing.
DeepSDF and ConvONet reply on latent feature space to decode geometries, which shows the potential in regularizing geometries.
However, the superiosity of our model suggests our dictionary based representation is advantageous over conditional implicit representation. 

\section{Related Work}

\paragraph{Generalizable Implicit Neural Representations.}
Implicit Neural Representation (INR) \cite{tancik2020fourier, sitzmann2020implicit} notoriously suffers from the limited cross-scene generalization capability.
\citet{tancik2021learned, sitzmann2020metasdf} propose meta-learning based algorithms to better initialize INR weights for fast convergence.
\citet{chen2021learning, park2019deepsdf, chabra2020deep, chibane2020implicit, jang2021codenerf, martin2021nerf, rematas2021sharf} introduce learnable latent embeddings to encode scene specific information and condition the INR on the latent code for generalizable representation. 
In \citet{sitzmann2020implicit}, the authors further utilize a hyper-network \citep{ha2016hypernetworks} to predict INR weights directly from inputs.
Compared with conditional fields or hyper-network based methods, sparse coding based NID, with just one last layer, can achieve faster adaptation.
The dictionary representation simplifies the mapping between latent spaces to a sparse linear combination over the additive basis, which can be manipulated more interpretably and also contributes to transferability. Last but not least, it is known that imposing sparsity can help overcome noise in ill-posed inverse problems \citep{donoho2006compressed, candes2011robust}.

\paragraph{Mixture of Experts (MoE).}
Mixture of Experts \citep{jacobs1991adaptive,jordan1994hierarchical,chen1999improved,6215056,roller2021hash} perform conditional computations composed of a group of parallel sub-models (a.k.a. experts) according to a routing policies \citep{dua2021tricks,roller2021hash}.
Recent advances \citep{shazeer2017outrageously,lepikhin2020gshard,fedus2021switch} improve MoE by adopting a sparse-gating strategy, which only activates a minority of experts by selecting top candidates according to the scores given by the gating networks.
This brings massive advantages in model capacity, training time, and achieved performance \citep{shazeer2017outrageously}.
\citet{fedus2021switch} even built language models with trillions of parameters.
To stabilize the training, \citet{hansen1999combining,lepikhin2020gshard,fedus2021switch} investigated auxiliary loading loss to balance the selection of experts. Alternatively, \citet{lewis2021base,clark2022unified} encourage a balanced routing by solving a linear assignment problem.


\section{Conclusion}

We propose Neural Implicit Dictionary (NID) learned from data collection to represent the signals as a sparse combination of the function basis inside.
Unlike tradition dictionary, our NID contains continuous function basis, which are parameterized by subnetworks.
To train thousands of networks efficiently, we employ Mixture-of-Expert training strategy.
Our NID enjoys higher compactness, robustness, and generalization.
Our experiments demonstrate promising applications of NID in instant regression, image inpainting, video decomposition, and reconstruction from sparse observations. Our future work may bring in subspace learning theories to analyze NID.

\section*{Acknowledgement}

Z. W. is in part supported by a US Army Research Office Young Investigator Award (W911NF2010240).

\nocite{langley00}

\bibliography{example_paper}

\begin{thebibliography}{77}
\providecommand{\natexlab}[1]{#1}
\providecommand{\url}[1]{\texttt{#1}}
\expandafter\ifx\csname urlstyle\endcsname\relax
  \providecommand{\doi}[1]{doi: #1}\else
  \providecommand{\doi}{doi: \begingroup \urlstyle{rm}\Url}\fi

\bibitem[Aharon et~al.(2006)Aharon, Elad, and Bruckstein]{aharon2006k}
Aharon, M., Elad, M., and Bruckstein, A.
\newblock K-svd: An algorithm for designing overcomplete dictionaries for
  sparse representation.
\newblock \emph{IEEE Transactions on signal processing}, 54\penalty0
  (11):\penalty0 4311--4322, 2006.

\bibitem[Attal et~al.(2021{\natexlab{a}})Attal, Huang, Zollhoefer, Kopf, and
  Kim]{attal2021learning}
Attal, B., Huang, J.-B., Zollhoefer, M., Kopf, J., and Kim, C.
\newblock Learning neural light fields with ray-space embedding networks.
\newblock \emph{arXiv preprint arXiv:2112.01523}, 2021{\natexlab{a}}.

\bibitem[Attal et~al.(2021{\natexlab{b}})Attal, Laidlaw, Gokaslan, Kim,
  Richardt, Tompkin, and O'Toole]{attal2021torf}
Attal, B., Laidlaw, E., Gokaslan, A., Kim, C., Richardt, C., Tompkin, J., and
  O'Toole, M.
\newblock T{\"o}rf: Time-of-flight radiance fields for dynamic scene view
  synthesis.
\newblock \emph{Advances in neural information processing systems}, 34,
  2021{\natexlab{b}}.

\bibitem[Bengio et~al.(2015)Bengio, Bacon, Pineau, and
  Precup]{bengio2015conditional}
Bengio, E., Bacon, P.-L., Pineau, J., and Precup, D.
\newblock Conditional computation in neural networks for faster models.
\newblock \emph{arXiv preprint arXiv:1511.06297}, 2015.

\bibitem[Bojanowski et~al.(2017)Bojanowski, Joulin, Lopez-Paz, and
  Szlam]{bojanowski2017optimizing}
Bojanowski, P., Joulin, A., Lopez-Paz, D., and Szlam, A.
\newblock Optimizing the latent space of generative networks.
\newblock \emph{arXiv preprint arXiv:1707.05776}, 2017.

\bibitem[Cand{\`e}s et~al.(2006)Cand{\`e}s, Romberg, and Tao]{candes2006robust}
Cand{\`e}s, E.~J., Romberg, J., and Tao, T.
\newblock Robust uncertainty principles: Exact signal reconstruction from
  highly incomplete frequency information.
\newblock \emph{IEEE Transactions on information theory}, 52\penalty0
  (2):\penalty0 489--509, 2006.

\bibitem[Cand{\`e}s et~al.(2011)Cand{\`e}s, Li, Ma, and
  Wright]{candes2011robust}
Cand{\`e}s, E.~J., Li, X., Ma, Y., and Wright, J.
\newblock Robust principal component analysis?
\newblock \emph{Journal of the ACM (JACM)}, 58\penalty0 (3):\penalty0 1--37,
  2011.

\bibitem[Chabra et~al.(2020)Chabra, Lenssen, Ilg, Schmidt, Straub, Lovegrove,
  and Newcombe]{chabra2020deep}
Chabra, R., Lenssen, J.~E., Ilg, E., Schmidt, T., Straub, J., Lovegrove, S.,
  and Newcombe, R.
\newblock Deep local shapes: Learning local sdf priors for detailed 3d
  reconstruction.
\newblock In \emph{European Conference on Computer Vision}, pp.\  608--625.
  Springer, 2020.

\bibitem[Chan et~al.(2015)Chan, Jia, Gao, Lu, Zeng, and Ma]{chan2015pcanet}
Chan, T.-H., Jia, K., Gao, S., Lu, J., Zeng, Z., and Ma, Y.
\newblock Pcanet: A simple deep learning baseline for image classification?
\newblock \emph{IEEE transactions on image processing}, 24\penalty0
  (12):\penalty0 5017--5032, 2015.

\bibitem[Chang et~al.(2015)Chang, Funkhouser, Guibas, Hanrahan, Huang, Li,
  Savarese, Savva, Song, Su, Xiao, Yi, and Yu]{shapenet2015}
Chang, A.~X., Funkhouser, T., Guibas, L., Hanrahan, P., Huang, Q., Li, Z.,
  Savarese, S., Savva, M., Song, S., Su, H., Xiao, J., Yi, L., and Yu, F.
\newblock {ShapeNet: An Information-Rich 3D Model Repository}.
\newblock Technical Report arXiv:1512.03012 [cs.GR], Stanford University ---
  Princeton University --- Toyota Technological Institute at Chicago, 2015.

\bibitem[Chen et~al.(2021{\natexlab{a}})Chen, Xu, Zhao, Zhang, Xiang, Yu, and
  Su]{chen2021mvsnerf}
Chen, A., Xu, Z., Zhao, F., Zhang, X., Xiang, F., Yu, J., and Su, H.
\newblock Mvsnerf: Fast generalizable radiance field reconstruction from
  multi-view stereo.
\newblock \emph{arXiv preprint arXiv:2103.15595}, 2021{\natexlab{a}}.

\bibitem[Chen \& Needell(2016)Chen and Needell]{chen2016compressed}
Chen, G. and Needell, D.
\newblock Compressed sensing and dictionary learning.
\newblock \emph{Finite Frame Theory: A Complete Introduction to
  Overcompleteness}, 73:\penalty0 201, 2016.

\bibitem[Chen et~al.(2021{\natexlab{b}})Chen, He, Wang, Ren, Lim, and
  Shrivastava]{chen2021nerv}
Chen, H., He, B., Wang, H., Ren, Y., Lim, S.~N., and Shrivastava, A.
\newblock Nerv: Neural representations for videos.
\newblock \emph{Advances in Neural Information Processing Systems}, 34,
  2021{\natexlab{b}}.

\bibitem[Chen et~al.(1999)Chen, Xu, and Chi]{chen1999improved}
Chen, K., Xu, L., and Chi, H.
\newblock Improved learning algorithms for mixture of experts in multiclass
  classification.
\newblock \emph{Neural networks}, 12\penalty0 (9):\penalty0 1229--1252, 1999.

\bibitem[Chen et~al.(2021{\natexlab{c}})Chen, Liu, and Wang]{chen2021learning}
Chen, Y., Liu, S., and Wang, X.
\newblock Learning continuous image representation with local implicit image
  function.
\newblock In \emph{Proceedings of the IEEE/CVF Conference on Computer Vision
  and Pattern Recognition}, pp.\  8628--8638, 2021{\natexlab{c}}.

\bibitem[Chibane et~al.(2020)Chibane, Alldieck, and
  Pons-Moll]{chibane2020implicit}
Chibane, J., Alldieck, T., and Pons-Moll, G.
\newblock Implicit functions in feature space for 3d shape reconstruction and
  completion.
\newblock In \emph{Proceedings of the IEEE/CVF Conference on Computer Vision
  and Pattern Recognition}, pp.\  6970--6981, 2020.

\bibitem[Choy et~al.(2016)Choy, Xu, Gwak, Chen, and Savarese]{choy20163d}
Choy, C.~B., Xu, D., Gwak, J., Chen, K., and Savarese, S.
\newblock 3d-r2n2: A unified approach for single and multi-view 3d object
  reconstruction.
\newblock In \emph{Proceedings of the European Conference on Computer Vision
  ({ECCV})}, 2016.

\bibitem[Clark et~al.(2022)Clark, Casas, Guy, Mensch, Paganini, Hoffmann,
  Damoc, Hechtman, Cai, Borgeaud, et~al.]{clark2022unified}
Clark, A., Casas, D. d.~l., Guy, A., Mensch, A., Paganini, M., Hoffmann, J.,
  Damoc, B., Hechtman, B., Cai, T., Borgeaud, S., et~al.
\newblock Unified scaling laws for routed language models.
\newblock \emph{arXiv preprint arXiv:2202.01169}, 2022.

\bibitem[Donoho(2006)]{donoho2006compressed}
Donoho, D.~L.
\newblock Compressed sensing.
\newblock \emph{IEEE Transactions on information theory}, 52\penalty0
  (4):\penalty0 1289--1306, 2006.

\bibitem[Dua et~al.(2021)Dua, Bhosale, Goswami, Cross, Lewis, and
  Fan]{dua2021tricks}
Dua, D., Bhosale, S., Goswami, V., Cross, J., Lewis, M., and Fan, A.
\newblock Tricks for training sparse translation models.
\newblock \emph{arXiv preprint arXiv:2110.08246}, 2021.

\bibitem[Fan et~al.(2018)Fan, Sun, Ding, Huang, Cai, and
  Paisley]{fan2018segmentation}
Fan, Z., Sun, L., Ding, X., Huang, Y., Cai, C., and Paisley, J.
\newblock A segmentation-aware deep fusion network for compressed sensing mri.
\newblock In \emph{Proceedings of the European Conference on Computer Vision
  (ECCV)}, pp.\  55--70, 2018.

\bibitem[Fan et~al.(2022)Fan, Jiang, Wang, Gong, Xu, and Wang]{fan2022unified}
Fan, Z., Jiang, Y., Wang, P., Gong, X., Xu, D., and Wang, Z.
\newblock Unified implicit neural stylization.
\newblock \emph{arXiv preprint arXiv:2204.01943}, 2022.

\bibitem[Fedus et~al.(2021)Fedus, Zoph, and Shazeer]{fedus2021switch}
Fedus, W., Zoph, B., and Shazeer, N.
\newblock Switch transformers: Scaling to trillion parameter models with simple
  and efficient sparsity.
\newblock \emph{arXiv preprint arXiv:2101.03961}, 2021.

\bibitem[Feng \& Varshney(2021)Feng and Varshney]{feng2021signet}
Feng, B.~Y. and Varshney, A.
\newblock Signet: Efficient neural representation for light fields.
\newblock In \emph{Proceedings of the IEEE/CVF International Conference on
  Computer Vision}, pp.\  14224--14233, 2021.

\bibitem[Gordon et~al.(2019)Gordon, Bruinsma, Foong, Requeima, Dubois, and
  Turner]{gordon2019convolutional}
Gordon, J., Bruinsma, W.~P., Foong, A.~Y., Requeima, J., Dubois, Y., and
  Turner, R.~E.
\newblock Convolutional conditional neural processes.
\newblock \emph{arXiv preprint arXiv:1910.13556}, 2019.

\bibitem[Gropp et~al.(2020)Gropp, Yariv, Haim, Atzmon, and
  Lipman]{gropp2020implicit}
Gropp, A., Yariv, L., Haim, N., Atzmon, M., and Lipman, Y.
\newblock Implicit geometric regularization for learning shapes.
\newblock \emph{arXiv preprint arXiv:2002.10099}, 2020.

\bibitem[Ha et~al.(2016)Ha, Dai, and Le]{ha2016hypernetworks}
Ha, D., Dai, A., and Le, Q.~V.
\newblock Hypernetworks.
\newblock \emph{arXiv preprint arXiv:1609.09106}, 2016.

\bibitem[Han et~al.(2018)Han, Jentzen, and Weinan]{han2018solving}
Han, J., Jentzen, A., and Weinan, E.
\newblock Solving high-dimensional partial differential equations using deep
  learning.
\newblock \emph{Proceedings of the National Academy of Sciences}, 115\penalty0
  (34):\penalty0 8505--8510, 2018.

\bibitem[Hansen(1999)]{hansen1999combining}
Hansen, J.~V.
\newblock Combining predictors: comparison of five meta machine learning
  methods.
\newblock \emph{Information Sciences}, 119\penalty0 (1-2):\penalty0 91--105,
  1999.

\bibitem[He et~al.(2021)He, Qiu, Zeng, Yang, Zhai, and Tang]{he2021fastmoe}
He, J., Qiu, J., Zeng, A., Yang, Z., Zhai, J., and Tang, J.
\newblock Fastmoe: A fast mixture-of-expert training system.
\newblock \emph{arXiv preprint arXiv:2103.13262}, 2021.

\bibitem[He et~al.(2016)He, Zhang, Ren, and Sun]{he2016deep}
He, K., Zhang, X., Ren, S., and Sun, J.
\newblock Deep residual learning for image recognition.
\newblock In \emph{CVPR}, pp.\  770--778, 2016.

\bibitem[Huang et~al.(2018)Huang, Su, and Guibas]{huang2018robust}
Huang, J., Su, H., and Guibas, L.
\newblock Robust watertight manifold surface generation method for shapenet
  models.
\newblock \emph{arXiv preprint arXiv:1802.01698}, 2018.

\bibitem[Jacobs et~al.(1991)Jacobs, Jordan, Nowlan, and
  Hinton]{jacobs1991adaptive}
Jacobs, R.~A., Jordan, M.~I., Nowlan, S.~J., and Hinton, G.~E.
\newblock Adaptive mixtures of local experts.
\newblock \emph{Neural computation}, 3\penalty0 (1):\penalty0 79--87, 1991.

\bibitem[Jang \& Agapito(2021)Jang and Agapito]{jang2021codenerf}
Jang, W. and Agapito, L.
\newblock Codenerf: Disentangled neural radiance fields for object categories.
\newblock In \emph{Proceedings of the IEEE/CVF International Conference on
  Computer Vision}, pp.\  12949--12958, 2021.

\bibitem[Ji et~al.(2010)Ji, Liu, Shen, and Xu]{ji2010robust}
Ji, H., Liu, C., Shen, Z., and Xu, Y.
\newblock Robust video denoising using low rank matrix completion.
\newblock In \emph{2010 IEEE Computer Society Conference on Computer Vision and
  Pattern Recognition}, pp.\  1791--1798. IEEE, 2010.

\bibitem[Jordan \& Jacobs(1994)Jordan and Jacobs]{jordan1994hierarchical}
Jordan, M.~I. and Jacobs, R.~A.
\newblock Hierarchical mixtures of experts and the em algorithm.
\newblock \emph{Neural computation}, 6\penalty0 (2):\penalty0 181--214, 1994.

\bibitem[Kreutz-Delgado et~al.(2003)Kreutz-Delgado, Murray, Rao, Engan, Lee,
  and Sejnowski]{kreutz2003dictionary}
Kreutz-Delgado, K., Murray, J.~F., Rao, B.~D., Engan, K., Lee, T.-W., and
  Sejnowski, T.~J.
\newblock Dictionary learning algorithms for sparse representation.
\newblock \emph{Neural computation}, 15\penalty0 (2):\penalty0 349--396, 2003.

\bibitem[Lepikhin et~al.(2020)Lepikhin, Lee, Xu, Chen, Firat, Huang, Krikun,
  Shazeer, and Chen]{lepikhin2020gshard}
Lepikhin, D., Lee, H., Xu, Y., Chen, D., Firat, O., Huang, Y., Krikun, M.,
  Shazeer, N., and Chen, Z.
\newblock Gshard: Scaling giant models with conditional computation and
  automatic sharding.
\newblock \emph{arXiv preprint arXiv:2006.16668}, 2020.

\bibitem[Lewis et~al.(2021)Lewis, Bhosale, Dettmers, Goyal, and
  Zettlemoyer]{lewis2021base}
Lewis, M., Bhosale, S., Dettmers, T., Goyal, N., and Zettlemoyer, L.
\newblock Base layers: Simplifying training of large, sparse models.
\newblock In \emph{International Conference on Machine Learning}, pp.\
  6265--6274. PMLR, 2021.

\bibitem[Li et~al.(2020)Li, Kovachki, Azizzadenesheli, Liu, Bhattacharya,
  Stuart, and Anandkumar]{li2020fourier}
Li, Z., Kovachki, N., Azizzadenesheli, K., Liu, B., Bhattacharya, K., Stuart,
  A., and Anandkumar, A.
\newblock Fourier neural operator for parametric partial differential
  equations.
\newblock \emph{arXiv preprint arXiv:2010.08895}, 2020.

\bibitem[Liu et~al.(2018)Liu, Reda, Shih, Wang, Tao, and
  Catanzaro]{liu2018image}
Liu, G., Reda, F.~A., Shih, K.~J., Wang, T.-C., Tao, A., and Catanzaro, B.
\newblock Image inpainting for irregular holes using partial convolutions.
\newblock In \emph{Proceedings of the European Conference on Computer Vision
  (ECCV)}, pp.\  85--100, 2018.

\bibitem[Liu et~al.(2015)Liu, Luo, Wang, and Tang]{liu2015faceattributes}
Liu, Z., Luo, P., Wang, X., and Tang, X.
\newblock Deep learning face attributes in the wild.
\newblock In \emph{Proceedings of International Conference on Computer Vision
  (ICCV)}, December 2015.

\bibitem[Lustig et~al.(2008)Lustig, Donoho, Santos, and
  Pauly]{lustig2008compressed}
Lustig, M., Donoho, D.~L., Santos, J.~M., and Pauly, J.~M.
\newblock Compressed sensing mri.
\newblock \emph{IEEE signal processing magazine}, 25\penalty0 (2):\penalty0
  72--82, 2008.

\bibitem[Martin-Brualla et~al.(2021)Martin-Brualla, Radwan, Sajjadi, Barron,
  Dosovitskiy, and Duckworth]{martin2021nerf}
Martin-Brualla, R., Radwan, N., Sajjadi, M.~S., Barron, J.~T., Dosovitskiy, A.,
  and Duckworth, D.
\newblock Nerf in the wild: Neural radiance fields for unconstrained photo
  collections.
\newblock In \emph{Proceedings of the IEEE/CVF Conference on Computer Vision
  and Pattern Recognition}, pp.\  7210--7219, 2021.

\bibitem[Mescheder et~al.(2019)Mescheder, Oechsle, Niemeyer, Nowozin, and
  Geiger]{mescheder2019occupancy}
Mescheder, L., Oechsle, M., Niemeyer, M., Nowozin, S., and Geiger, A.
\newblock Occupancy networks: Learning 3d reconstruction in function space.
\newblock In \emph{Proceedings of the IEEE/CVF Conference on Computer Vision
  and Pattern Recognition}, pp.\  4460--4470, 2019.

\bibitem[Metzler et~al.(2016)Metzler, Maleki, and
  Baraniuk]{metzler2016denoising}
Metzler, C.~A., Maleki, A., and Baraniuk, R.~G.
\newblock From denoising to compressed sensing.
\newblock \emph{IEEE Transactions on Information Theory}, 62\penalty0
  (9):\penalty0 5117--5144, 2016.

\bibitem[Mildenhall et~al.(2020)Mildenhall, Srinivasan, Tancik, Barron,
  Ramamoorthi, and Ng]{mildenhall2020nerf}
Mildenhall, B., Srinivasan, P.~P., Tancik, M., Barron, J.~T., Ramamoorthi, R.,
  and Ng, R.
\newblock Nerf: Representing scenes as neural radiance fields for view
  synthesis.
\newblock In \emph{European conference on computer vision}, pp.\  405--421.
  Springer, 2020.

\bibitem[Papyan et~al.(2017)Papyan, Romano, and Elad]{papyan2017convolutional}
Papyan, V., Romano, Y., and Elad, M.
\newblock Convolutional neural networks analyzed via convolutional sparse
  coding.
\newblock \emph{The Journal of Machine Learning Research}, 18\penalty0
  (1):\penalty0 2887--2938, 2017.

\bibitem[Park et~al.(2019)Park, Florence, Straub, Newcombe, and
  Lovegrove]{park2019deepsdf}
Park, J.~J., Florence, P., Straub, J., Newcombe, R., and Lovegrove, S.
\newblock Deepsdf: Learning continuous signed distance functions for shape
  representation.
\newblock In \emph{Proceedings of the IEEE/CVF Conference on Computer Vision
  and Pattern Recognition}, pp.\  165--174, 2019.

\bibitem[Peng et~al.(2020)Peng, Niemeyer, Mescheder, Pollefeys, and
  Geiger]{peng2020convolutional}
Peng, S., Niemeyer, M., Mescheder, L., Pollefeys, M., and Geiger, A.
\newblock Convolutional occupancy networks.
\newblock In \emph{Computer Vision--ECCV 2020: 16th European Conference,
  Glasgow, UK, August 23--28, 2020, Proceedings, Part III 16}, pp.\  523--540.
  Springer, 2020.

\bibitem[Qi et~al.(2017{\natexlab{a}})Qi, Su, Mo, and Guibas]{qi2017pointnet}
Qi, C.~R., Su, H., Mo, K., and Guibas, L.~J.
\newblock Pointnet: Deep learning on point sets for 3d classification and
  segmentation.
\newblock In \emph{Proceedings of the IEEE conference on computer vision and
  pattern recognition}, pp.\  652--660, 2017{\natexlab{a}}.

\bibitem[Qi et~al.(2017{\natexlab{b}})Qi, Yi, Su, and Guibas]{qi2017pointnet++}
Qi, C.~R., Yi, L., Su, H., and Guibas, L.~J.
\newblock Pointnet++: Deep hierarchical feature learning on point sets in a
  metric space.
\newblock \emph{arXiv preprint arXiv:1706.02413}, 2017{\natexlab{b}}.

\bibitem[Reiser et~al.(2021)Reiser, Peng, Liao, and Geiger]{reiser2021kilonerf}
Reiser, C., Peng, S., Liao, Y., and Geiger, A.
\newblock Kilonerf: Speeding up neural radiance fields with thousands of tiny
  mlps.
\newblock \emph{arXiv preprint arXiv:2103.13744}, 2021.

\bibitem[Rematas et~al.(2021)Rematas, Martin-Brualla, and
  Ferrari]{rematas2021sharf}
Rematas, K., Martin-Brualla, R., and Ferrari, V.
\newblock Sharf: Shape-conditioned radiance fields from a single view.
\newblock \emph{arXiv preprint arXiv:2102.08860}, 2021.

\bibitem[Roller et~al.(2021)Roller, Sukhbaatar, Szlam, and
  Weston]{roller2021hash}
Roller, S., Sukhbaatar, S., Szlam, A., and Weston, J.~E.
\newblock Hash layers for large sparse models.
\newblock In Beygelzimer, A., Dauphin, Y., Liang, P., and Vaughan, J.~W.
  (eds.), \emph{Advances in Neural Information Processing Systems}, 2021.
\newblock URL \url{https://openreview.net/forum?id=lMgDDWb1ULW}.

\bibitem[Saito et~al.(2019)Saito, Huang, Natsume, Morishima, Kanazawa, and
  Li]{saito2019pifu}
Saito, S., Huang, Z., Natsume, R., Morishima, S., Kanazawa, A., and Li, H.
\newblock Pifu: Pixel-aligned implicit function for high-resolution clothed
  human digitization.
\newblock In \emph{Proceedings of the IEEE/CVF International Conference on
  Computer Vision}, pp.\  2304--2314, 2019.

\bibitem[Shazeer et~al.(2017)Shazeer, Mirhoseini, Maziarz, Davis, Le, Hinton,
  and Dean]{shazeer2017outrageously}
Shazeer, N., Mirhoseini, A., Maziarz, K., Davis, A., Le, Q., Hinton, G., and
  Dean, J.
\newblock Outrageously large neural networks: The sparsely-gated
  mixture-of-experts layer.
\newblock \emph{arXiv preprint arXiv:1701.06538}, 2017.

\bibitem[Shen et~al.(2021)Shen, Wang, Liu, Pan, Li, Gao, Li, and
  Yu]{shen2021non}
Shen, S., Wang, Z., Liu, P., Pan, Z., Li, R., Gao, T., Li, S., and Yu, J.
\newblock Non-line-of-sight imaging via neural transient fields.
\newblock \emph{IEEE Transactions on Pattern Analysis and Machine
  Intelligence}, 2021.

\bibitem[Shepp \& Logan(1974)Shepp and Logan]{shepp1974fourier}
Shepp, L.~A. and Logan, B.~F.
\newblock The fourier reconstruction of a head section.
\newblock \emph{IEEE Transactions on nuclear science}, 21\penalty0
  (3):\penalty0 21--43, 1974.

\bibitem[Sitzmann et~al.(2020{\natexlab{a}})Sitzmann, Chan, Tucker, Snavely,
  and Wetzstein]{sitzmann2020metasdf}
Sitzmann, V., Chan, E.~R., Tucker, R., Snavely, N., and Wetzstein, G.
\newblock Metasdf: Meta-learning signed distance functions.
\newblock \emph{arXiv preprint arXiv:2006.09662}, 2020{\natexlab{a}}.

\bibitem[Sitzmann et~al.(2020{\natexlab{b}})Sitzmann, Martel, Bergman, Lindell,
  and Wetzstein]{sitzmann2020implicit}
Sitzmann, V., Martel, J., Bergman, A., Lindell, D., and Wetzstein, G.
\newblock Implicit neural representations with periodic activation functions.
\newblock \emph{Advances in Neural Information Processing Systems}, 33,
  2020{\natexlab{b}}.

\bibitem[Sitzmann et~al.(2021)Sitzmann, Rezchikov, Freeman, Tenenbaum, and
  Durand]{sitzmann2021light}
Sitzmann, V., Rezchikov, S., Freeman, W.~T., Tenenbaum, J.~B., and Durand, F.
\newblock Light field networks: Neural scene representations with
  single-evaluation rendering.
\newblock \emph{arXiv preprint arXiv:2106.02634}, 2021.

\bibitem[Sutherland \& Schneider(2015)Sutherland and
  Schneider]{sutherland2015error}
Sutherland, D.~J. and Schneider, J.
\newblock On the error of random fourier features.
\newblock \emph{arXiv preprint arXiv:1506.02785}, 2015.

\bibitem[Tancik et~al.(2020)Tancik, Srinivasan, Mildenhall, Fridovich-Keil,
  Raghavan, Singhal, Ramamoorthi, Barron, and Ng]{tancik2020fourier}
Tancik, M., Srinivasan, P.~P., Mildenhall, B., Fridovich-Keil, S., Raghavan,
  N., Singhal, U., Ramamoorthi, R., Barron, J.~T., and Ng, R.
\newblock Fourier features let networks learn high frequency functions in low
  dimensional domains.
\newblock \emph{arXiv preprint arXiv:2006.10739}, 2020.

\bibitem[Tancik et~al.(2021)Tancik, Mildenhall, Wang, Schmidt, Srinivasan,
  Barron, and Ng]{tancik2021learned}
Tancik, M., Mildenhall, B., Wang, T., Schmidt, D., Srinivasan, P.~P., Barron,
  J.~T., and Ng, R.
\newblock Learned initializations for optimizing coordinate-based neural
  representations.
\newblock In \emph{Proceedings of the IEEE/CVF Conference on Computer Vision
  and Pattern Recognition}, pp.\  2846--2855, 2021.

\bibitem[Tariyal et~al.(2016)Tariyal, Majumdar, Singh, and
  Vatsa]{tariyal2016deep}
Tariyal, S., Majumdar, A., Singh, R., and Vatsa, M.
\newblock Deep dictionary learning.
\newblock \emph{IEEE Access}, 4:\penalty0 10096--10109, 2016.

\bibitem[To{\v{s}}i{\'c} \& Frossard(2011)To{\v{s}}i{\'c} and
  Frossard]{tovsic2011dictionary}
To{\v{s}}i{\'c}, I. and Frossard, P.
\newblock Dictionary learning.
\newblock \emph{IEEE Signal Processing Magazine}, 28\penalty0 (2):\penalty0
  27--38, 2011.

\bibitem[Turki et~al.(2021)Turki, Ramanan, and Satyanarayanan]{turki2021mega}
Turki, H., Ramanan, D., and Satyanarayanan, M.
\newblock Mega-nerf: Scalable construction of large-scale nerfs for virtual
  fly-throughs.
\newblock \emph{arXiv preprint arXiv:2112.10703}, 2021.

\bibitem[Vacavant et~al.(2012)Vacavant, Chateau, Wilhelm, and
  Lequievre]{vacavant2012benchmark}
Vacavant, A., Chateau, T., Wilhelm, A., and Lequievre, L.
\newblock A benchmark dataset for outdoor foreground/background extraction.
\newblock In \emph{Asian Conference on Computer Vision}, pp.\  291--300.
  Springer, 2012.

\bibitem[Wang et~al.(2021)Wang, Wang, Genova, Srinivasan, Zhou, Barron,
  Martin-Brualla, Snavely, and Funkhouser]{wang2021ibrnet}
Wang, Q., Wang, Z., Genova, K., Srinivasan, P.~P., Zhou, H., Barron, J.~T.,
  Martin-Brualla, R., Snavely, N., and Funkhouser, T.
\newblock Ibrnet: Learning multi-view image-based rendering.
\newblock In \emph{Proceedings of the IEEE/CVF Conference on Computer Vision
  and Pattern Recognition}, pp.\  4690--4699, 2021.

\bibitem[Wang et~al.(2004)Wang, Bovik, Sheikh, and Simoncelli]{wang2004image}
Wang, Z., Bovik, A.~C., Sheikh, H.~R., and Simoncelli, E.~P.
\newblock Image quality assessment: from error visibility to structural
  similarity.
\newblock \emph{IEEE transactions on image processing}, 13\penalty0
  (4):\penalty0 600--612, 2004.

\bibitem[Yu et~al.(2021)Yu, Ye, Tancik, and Kanazawa]{yu2021pixelnerf}
Yu, A., Ye, V., Tancik, M., and Kanazawa, A.
\newblock pixelnerf: Neural radiance fields from one or few images.
\newblock In \emph{Proceedings of the IEEE/CVF Conference on Computer Vision
  and Pattern Recognition}, pp.\  4578--4587, 2021.

\bibitem[Yu et~al.(2019)Yu, Lin, Yang, Shen, Lu, and Huang]{yu2019free}
Yu, J., Lin, Z., Yang, J., Shen, X., Lu, X., and Huang, T.~S.
\newblock Free-form image inpainting with gated convolution.
\newblock In \emph{Proceedings of the IEEE/CVF International Conference on
  Computer Vision}, pp.\  4471--4480, 2019.

\bibitem[Yuksel et~al.(2012)Yuksel, Wilson, and Gader]{6215056}
Yuksel, S.~E., Wilson, J.~N., and Gader, P.~D.
\newblock Twenty years of mixture of experts.
\newblock \emph{IEEE Transactions on Neural Networks and Learning Systems},
  23\penalty0 (8):\penalty0 1177--1193, 2012.
\newblock \doi{10.1109/TNNLS.2012.2200299}.

\bibitem[Zaheer et~al.(2017)Zaheer, Kottur, Ravanbakhsh, Poczos, Salakhutdinov,
  and Smola]{zaheer2017deep}
Zaheer, M., Kottur, S., Ravanbakhsh, S., Poczos, B., Salakhutdinov, R., and
  Smola, A.
\newblock Deep sets.
\newblock \emph{arXiv preprint arXiv:1703.06114}, 2017.

\bibitem[Zhang et~al.(2018)Zhang, Isola, Efros, Shechtman, and
  Wang]{zhang2018unreasonable}
Zhang, R., Isola, P., Efros, A.~A., Shechtman, E., and Wang, O.
\newblock The unreasonable effectiveness of deep features as a perceptual
  metric.
\newblock In \emph{Proceedings of the IEEE conference on computer vision and
  pattern recognition}, pp.\  586--595, 2018.

\bibitem[Zhong et~al.(2021)Zhong, Bepler, Berger, and Davis]{zhong2021cryodrgn}
Zhong, E.~D., Bepler, T., Berger, B., and Davis, J.~H.
\newblock Cryodrgn: reconstruction of heterogeneous cryo-em structures using
  neural networks.
\newblock \emph{Nature Methods}, 18\penalty0 (2):\penalty0 176--185, 2021.

\end{thebibliography}
\bibliographystyle{icml2022}



\end{document}